\documentclass[letterpaper]{article} % DO NOT CHANGE THIS
\usepackage[preprint]{aaai2027}  % DO NOT CHANGE THIS
\usepackage[hyphens]{url}  % DO NOT CHANGE THIS
\usepackage{graphicx} % DO NOT CHANGE THIS
\usepackage{natbib}  % DO NOT CHANGE THIS AND DO NOT ADD ANY OPTIONS TO IT
\usepackage{caption} % DO NOT CHANGE THIS AND DO NOT ADD ANY OPTIONS TO IT
\usepackage{algorithm}
\usepackage{algorithmic}

\usepackage{booktabs}
\usepackage{tabularx}
\usepackage{array}

\usepackage{xcolor}
\usepackage{tikz}

\usepackage{newfloat}
\usepackage{listings}
\DeclareCaptionStyle{ruled}{labelfont=normalfont,labelsep=colon,strut=off} % DO NOT CHANGE THIS
\floatstyle{ruled}
\newfloat{listing}{tb}{lst}{}
\floatname{listing}{Listing}

\usepackage{booktabs}
\usepackage[table]{xcolor}

\definecolor{hanred}{RGB}{190,30,45}
\definecolor{dfblue}{RGB}{226,236,247}
\definecolor{dflight}{RGB}{241,245,250}

\newcommand{\qcell}[1]{\cellcolor{dfblue}#1}
\newcommand{\qcelllight}[1]{\cellcolor{dflight}#1}

\usepackage{amsmath}

\definecolor{flowbg}{HTML}{FFFFFF}
\definecolor{flowborder}{HTML}{000000}

\definecolor{flowblue}{HTML}{D0F3F8}    % cyan highlight
\definecolor{floworange}{HTML}{FFF8CC}  % yellow highlight
\definecolor{flowdiag}{HTML}{FFC8C9}    % pink highlight

\newcommand{\mS}{\mathbf{S}}

\newcommand{\hblue}[1]{%
  \begingroup
  \setlength{\fboxsep}{1.2pt}%
  \colorbox{flowblue}{\ensuremath{\displaystyle #1}}%
  \endgroup
}

\newcommand{\horange}[1]{%
  \begingroup
  \setlength{\fboxsep}{1.2pt}%
  \colorbox{floworange}{\ensuremath{\displaystyle #1}}%
  \endgroup
}

\newsavebox{\flowbox}

\usepackage{multirow}

\title{DeltaFlow: Noise-Adaptive Bidirectional Gated Delta Networks \\for Embedded Language Flows}

\author{
Guangfu Guo\textsuperscript{\rm 1},
Xiaoqian Lu\textsuperscript{\rm 2},
Linsey Pang\textsuperscript{\rm 3},
Weiran Yao\textsuperscript{\rm 4},\\
Haolin Chen\textsuperscript{\rm 4},
Kunpeng Liu\textsuperscript{\rm 1}\corresponding,
Long Cheng\textsuperscript{\rm 1}\corresponding
}

\affiliations{
\textsuperscript{\rm 1}Clemson University
\quad
\textsuperscript{\rm 2}University of Ottawa
\quad
\textsuperscript{\rm 3}PayPal
\quad
\textsuperscript{\rm 4}actAVA AI\\
gguo@clemson.edu, xlu053@uottawa.ca, panglinsey@gmail.com, weiran@actava.ai,\\ haolin@actava.ai, kunpenl@clemson.edu, lcheng2@clemson.edu

}

\begin{document}

\maketitle

\begin{abstract}
% Embedded Language Flows(ELF) perform iterative non-causal denoising in continuous token-latent space, but their full-attention backbone incurs quadratic sequence-mixing cost at every sampling step. Although Gated Delta Networks provide efficient recurrent sequence mixing, their causal scan and noise-independent memory updates are incompatible with bidirectional diffusion denoising.
% We introduce DeltaFlow, a prefix-preserving bidirectional GDN backbone that replaces most full-attention layers while retaining periodic dense-attention correction. DeltaFlow conditions recurrent decay and write rates on diffusion time and applies a scheduled representation-alignment objective to stabilize hidden states across neighboring noise levels. On OpenWebText under the SDE32 evaluation protocol, DeltaFlow-P achieves a generated perplexity of 21.228 at an entropy of 5.084, compared with 24.218 and 5.069 for ELF-full, while using 36B rather than 45B nominal training tokens. DeltaFlow-P achieves up to a $2.72\times$ speedup over ELF-full at a sequence length of 16k.

Embedded Language Flows (ELF) rely primarily on full non-causal attention for iterative denoising, repeatedly incurring quadratic sequence-mixing cost at each sampling step. Gated Delta Networks (GDNs) provide an efficient recurrent alternative, but their standard causal formulation cannot directly capture the bidirectional context required by ELF. We introduce DeltaFlow, a noise-adaptive bidirectional GDN backbone for continuous language denoising.  We study two variants: DeltaFlow-A, which alternates scan directions across layers, and DeltaFlow-P, which performs parallel forward and backward scans within each layer. We further introduce noise-adaptive memory control and scheduled Temporal State Consistency (TSC) to stabilize hidden representations across nearby noise levels. On OpenWebText, using a 32-step stochastic differential equation sampler, DeltaFlow-P reduces generated perplexity from 24.218 for the full-attention ELF baseline to 21.228 while maintaining comparable unigram entropy, with $36B$ training-token exposure compared with $45B$ for the baseline. In a denoiser-only benchmark, DeltaFlow-P achieves a 2.72× throughput speedup over the full-attention baseline at a sequence length of $16k$. These results show that DeltaFlow is a promising alternative to dense attention for efficient continuous language denoising.

% Embedded Language Flows (ELF) have recently shown promise for continuous language generation, but their reliance on full attention leads to repeated quadratic-cost sequence mixing during iterative denoising, limiting efficiency on long sequences. Gated Delta Networks (GDNs) provide an efficient subquadratic alternative; however, their standard causal recurrent formulation is not directly compatible with the non-causal denoising process required by ELF. In this paper, we propose DeltaFlow, a noise-adaptive bidirectional GDN backbone tailored for embedded language flow modeling. We provide two bidirectional GDN-based variants: an alternating-scan version and a  chunk-parallel version. To further adapt recurrent memory across diffusion timesteps, we introduce Noise-Adaptive Memory Control and Temporal State Consistenc, which mitigate state mismatch under different noise levels and stabilize hidden representations during denoising. Experiments show that, with 36B training tokens—less than the 45B tokens used by the full-attention ELF baseline—DeltaFlow reduces generation perplexity from 24.218 to 21.228 while maintaining a comparable entropy level of 5.084 (ours) versus 5.069. Code will be released soon.

\end{abstract}

\begin{figure*}[t]
\centering
\includegraphics[width=0.9\textwidth]{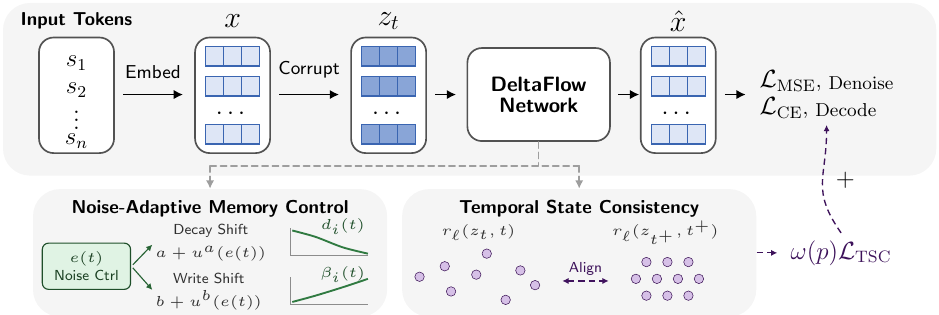}
% \caption{DeltaFlow training flow. Noise-adaptive memory control modulates GDN updates across timesteps, while Temporal State Consistenc (TSC) aligns representations at nearby noise levels.
% }
\caption{
\textbf{DeltaFlow training framework.}
DeltaFlow extends ELF with three components:
(1) prefix-preserving bidirectional GDN that reverses only target tokens;
(2) noise-adaptive memory control conditioned on diffusion time; and
(3) scheduled Temporal State Consistency (TSC) across neighboring noise levels.
}
%     DeltaFlow training flow with timestep-conditioned GDN memory
% control and TSC between nearby noise levels.
\label{fig:training_flow}
\end{figure*}

\section{Introduction}

Embedded Language Flows (ELF) generate text by iteratively denoising
continuous token latents~\citep{elf2026}. At each denoising step, every
target position may depend on both left and right contexts, together with
the diffusion timestep and self-conditioning signals; the denoising
backbone is therefore inherently non-causal. Existing ELF models rely
primarily on full attention for sequence mixing. For a sequence of length
$L$ and a sampler with $K$ denoising steps, this results in
$\mathcal{O}(KL^2)$ sequence-mixing cost, repeatedly paying for dense
token interactions throughout generation~\citep{vaswani2017attention}.
The resulting overhead becomes a major bottleneck as either sequence
length or sampling depth grows.

Linear-attention offer a more efficient alternative
by replacing dense pairwise interactions with compact states
~\citep{katharopoulos2020linear,schlag2021linear,sun2023retnet,
gu2024mamba,yang-etal-2024-gla}. Gated Delta Networks (GDNs) are
particularly attractive because they combine recurrent execution with
gated memory decay and residual delta updates~\citep{yang2025gatedgdn}.
However, two properties prevent a direct substitution of GDNs for ELF
attention layers. First, standard GDNs use a unidirectional causal scan,
whereas ELF requires bidirectional target context while preserving the
semantic order of prefix and control tokens that encode timestep, mode,
conditioning, and self-conditioning. Reversing the complete input would
therefore violate the ELF interface. Second, standard GDN decay and write
mechanisms have no direct diffusion-time conditioning, even though the
recurrent state is exposed to representations with substantially
different noise levels during denoising~\citep{ho2020denoising,elf2026}.
Adapting GDNs to ELF thus requires both prefix-preserving bidirectionality
and noise-aware memory control.

We introduce \textbf{DeltaFlow}, a family of bidirectional GDN denoisers
for Embedded Language Flows. DeltaFlow keeps prefix and control tokens in
their original order and reverses only the target-token body to construct
aligned forward and backward recurrent views. We develop two
realizations. \textbf{DeltaFlow-A} (\emph{Alternating-Scan}) alternates
scan directions across successive layers, whereas \textbf{DeltaFlow-P}
(\emph{Parallel-Scan}) performs and merges forward and reverse-body scans
within each GDN layer~\citep{afzal2026linear}. Both variants use a hybrid
stack that replaces most full-attention mixers with recurrent or
chunkwise GDN blocks while retaining periodic full non-causal attention
for direct token-level interaction~\citep{dig2025}. Consequently, most
sequence-mixing layers have linear or chunkwise-linear dependence on
$L$, although the complete hybrid denoiser and end-to-end generation
process are not strictly linear. DeltaFlow further introduces \emph{Noise-Adaptive Memory Control},
  which conditions GDN decay and write rates on diffusion time, together
  with scheduled \emph{Temporal State Consistency} (TSC), which aligns
  intermediate representations at neighboring noise levels along a
  shared-noise trajectory. Figure~\ref{fig:training_flow} summarizes the resulting
training framework.

We evaluate DeltaFlow under the generation and efficiency protocols used
by ELF. On OpenWebText, under the 1,000-sample SDE32 evaluation protocol
~\citep{song2021scorebased,elf2026}, DeltaFlow-P achieves a generated
perplexity of $21.228$ at a unigram entropy of $5.084$, compared with
$24.218$ and $5.069$ for the full-attention ELF baseline. This result is
obtained with $36$B nominal training-token exposure, compared with $45$B
for ELF-full. In a denoiser-only benchmark on a single H100,
DeltaFlow-P provides a $2.72\times$ throughput speedup at sequence length
$16$k. We further study the two bidirectional designs, the effects of
noise-adaptive memory control and TSC, transfer to conditional
generation, and computational efficiency across sequence lengths.

The main contributions of this paper are threefold:
\begin{itemize}
    \item We adapt causal GDNs to non-causal ELF denoising through a
    prefix-preserving bidirectional construction, with alternating-scan
    and parallel-scan realizations in a hybrid attention
    backbone.

    \item We introduce diffusion-time-conditioned decay and write
    control, together with scheduled TSC for stabilizing same-layer
    hidden representations across nearby noise levels.

    \item We provide a systematic evaluation of generation quality,
    diversity, conditional transfer, and long-sequence efficiency,
    demonstrating an improved perplexity--diversity trade-off and higher
    denoiser.
\end{itemize}

% \lin{maybe we move some content in introduction section to related work section...?  }

\section{Related Work}

\paragraph{Continuous Diffusion Language Modeling.}
Diffusion language models generate text through iterative denoising rather than autoregressive decoding. Diffusion-LM operates on continuous word embeddings~\citep{li2022diffusionlm}, SSD-LM diffuses vocabulary-simplex representations~\citep{han2023ssdlm}, and Embedded Language Flows (ELF) perform flow matching in continuous token-embedding space with endpoint prediction, self-conditioning, and an auxiliary decoding objective~\citep{elf2026}. Despite their different representations and training objectives, these methods largely rely on dense non-causal attention. DeltaFlow instead focuses on improving the efficiency of the ELF denoiser while retaining its latent-space and sampling interfaces, and adds TSC as an auxiliary denoising objective. TSC is related to cross-time output consistency in consistency models~\citep{song2023consistency} and noisy-to-clean feature alignment in REPA~\citep{yu2025repa}, but differs by aligning intermediate states of the same denoiser along a shared-noise path, without targeting few-step sampling or using an external encoder.

\paragraph{Efficient Recurrent and Linear Sequence Models.}
Linear-attention, recurrent, and state-space models replace dense pairwise interactions with compact hidden states. RetNet supports parallel and recurrent computation through retention~\citep{sun2023retnet}, Mamba introduces selective state updates~\citep{gu2024mamba}, and Gated Linear Attention uses data-dependent decay gates~\citep{yang-etal-2024-gla}. Gated Delta Networks further combine gated decay with residual delta updates~\citep{yang2025gatedgdn}. Recent GDN variants further improve the control of memory erasing and writing~\citep{hatamizadeh2026gdn2,sun2026fg2gdn}. However, their standard language-modeling formulations are causal and do not directly provide the bidirectional context required by ELF denoising.

\paragraph{Bidirectional and Diffusion-Oriented Efficient Backbones.}
Recent studies extend efficient sequence mixers beyond causal modeling. LION develops bidirectional recurrent and chunkwise formulations for linear-attention and DeltaNet-style models~\citep{afzal2026linear}, while DiG and SANA-WM combine efficient gated sequence mixing with diffusion backbones for image or video generation~\citep{dig2025,zhu2026sanawm}. DeltaFlow addresses a different setting: it preserves the semantic order of ELF prefix and control tokens, reverses only target tokens for bidirectional recurrence, and conditions GDN memory updates on diffusion time. It therefore provides an efficient non-causal backbone without modifying ELF's endpoint prediction, conditioning, decoding, or sampling procedures.

\section{Method}
\label{sec:method}

\paragraph{Method Overview.}
Figure~\ref{fig:training_flow} summarizes the DeltaFlow training
framework. The upper path preserves the original ELF latent-space
training pipeline. DeltaFlow introduces three modifications to this framework: a prefix-preserving bidirectional GDN
  sequence mixer, Noise-Adaptive Memory Control for timestep-conditioned modulation of recurrent
  decay and write rates, and Temporal State Consistency (TSC) across neighboring noise levels.

\subsection{Preserved ELF Interface}
\label{sec:elf-interface}

We briefly define the ELF interface required to describe DeltaFlow.
Let \(\mathbf{s}=(s_1,\ldots,s_L)\) be a token sequence, and let the
frozen ELF encoder produce its clean token-latent representation
{\small
\begin{equation}
    \mathbf{z}^{\star}
    =
    E_{\mathrm{ELF}}(\mathbf{s}).
    \label{eq:clean-latent}
\end{equation}
}
Let \(\Omega\) denote the body-token positions, excluding prefix and
padding positions. All token-level objectives are evaluated only over
\(\Omega\).

For the denoising branch, a diffusion time \(t\in[0,1]\) is sampled,
and each body-token latent is corrupted as
{\small
\begin{equation}
    \mathbf{z}_{t,i}
    =
    t\mathbf{z}^{\star}_{i}
    +
    (1-t)\boldsymbol{\epsilon}_{i},
    \qquad
    \boldsymbol{\epsilon}_{i}
    \sim
    \mathcal{N}
    \left(
        \mathbf{0},
        \sigma_{\mathrm{den}}^{2}\mathbf{I}
    \right),
    \quad i\in\Omega .
    \label{eq:elf-corruption}
\end{equation}
}Under our convention, \(t=0\) denotes the maximally corrupted endpoint and \(t=1\) denotes the clean-data endpoint; thus, larger \(t\) corresponds to a lower noise level.

Prefix and control tokens encode the diffusion timestep, generation
mode, conditioning information, and self-conditioning signal. They are
prepended to the noisy body latents but are not themselves denoising
targets. The shared backbone predicts the clean latent endpoint:
{\small
\begin{equation}
    \widehat{\mathbf{z}}^{\star}_{\theta,\Omega}
    =
    f_{\theta}
    \left(
        \left[
            \mathbf{p}
            \left(
                t,
                \mathrm{denoise},
                \widehat{\mathbf{z}}_{\mathrm{sc}}
            \right);
            \mathbf{z}_{t,\Omega}
        \right]
    \right)_{\Omega},
    \label{eq:elf-denoiser}
\end{equation}
}
where \(f_{\theta}\) denotes the shared latent backbone and
\(\mathbf{p}(\cdot)\) denotes the ELF prefix and control interface.

As shown in the upper path of Figure~\ref{fig:training_flow}, ELF
alternates between two training branches. The denoising branch uses the
velocity-based mean-squared-error objective
\(\mathcal{L}_{\mathrm{MSE}}\), whereas the decode branch uses the
token-level cross-entropy objective
\(\mathcal{L}_{\mathrm{CE}}\). Let
\(m\in\{\mathrm{denoise},\mathrm{decode}\}\) denote the sampled training
mode, with \(\Pr(m=\mathrm{decode})=\rho\). The preserved branch-specific base
objective is
{\small
\begin{equation}
    \mathcal{L}_{\mathrm{base}}(m)
    =
    \begin{cases}
        \mathcal{L}_{\mathrm{MSE}},
        & m=\mathrm{denoise},\\[2pt]
        \mathcal{L}_{\mathrm{CE}},
        & m=\mathrm{decode}.
    \end{cases}
    \label{eq:base-branch-loss}
\end{equation}
}
DeltaFlow modifies the sequence-mixing operations inside \(f_{\theta}\).
We next introduce the prefix-preserving bidirectional GDN that replaces
most full-attention mixers.

\subsection{Bidirectional GDN-Block}
\label{sec:bidirectional-gdn}

\begin{figure*}[t]
\centering
\includegraphics[width=0.8\textwidth]{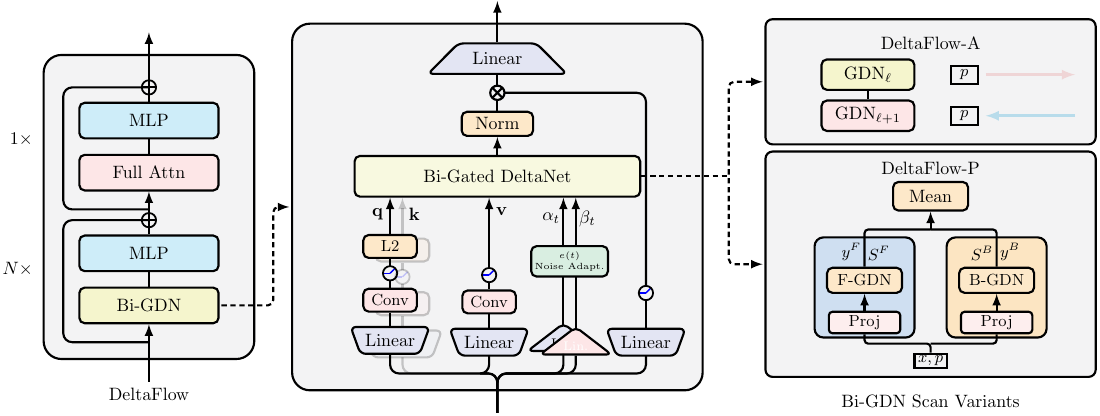}
% \caption{Overview of DeltaFlow. DeltaFlow combines prefix-preserving bidirectional GDN blocks, noise-adaptive memory control, and scheduled representation alignment for efficient non-causal denoising. 
% }
\caption{
\textbf{DeltaFlow architecture overview.}
\emph{Left:} A hybrid denoiser combining bidirectional GDN blocks with periodic full-attention blocks.
\emph{Center:} A bidirectional GDN block with noise-adaptive memory control.
\emph{Right:} DeltaFlow-A alternates scan directions across layers, whereas DeltaFlow-P performs and merges both directions within each layer.
}
\label{fig:method}
\end{figure*}

Figure~\ref{fig:method} illustrates the DeltaFlow architecture at three
levels. The left panel shows the hybrid denoiser stack, the center panel
details one bidirectional GDN block with noise-adaptive memory control,
and the right panel compares the Alternating-Scan and Parallel-Scan
variants. We first define the recurrent GDN operation and then describe
the two prefix-preserving bidirectional realizations.

\paragraph{Full-Attention Reference.}
Let $X=(p_1,\ldots,p_P,x_1,\ldots,x_L)$ be the prefix-plus-body sequence, where prefix tokens carry timestep, mode, and self-conditioning information. A full non-causal attention layer gives each target token a dense read over all prefix and target tokens:
{\small
\begin{align}
    \mathbf{y}_i^{\mathrm{full}}
    &=
    \sum_{j=1}^{P+L}
    \operatorname{softmax}_{j}
    \!\left(\mathbf{q}_i^\top[\mathbf{k}_1,\ldots,\mathbf{k}_{P+L}]\right)
    \mathbf{v}_j .
    \label{eq:full_attention_reference}
\end{align}
}
This is the natural ELF denoising mixer because target tokens are updated in parallel and may use both left and right context. Its cost is quadratic at every denoising step, so DeltaFlow keeps this support pattern but replaces most dense attention blocks with recurrent GDN memory.

\paragraph{From Attention to GDN Memory.}
Following the linear-attention view used by efficient diffusion backbones such as DiG~\citep{dig2025}, replacing the softmax kernel by a feature map $\phi(\cdot)$ turns the attention numerator and normalizer into recurrent states:

{\small
\begin{equation}
\mathbf{y}_i^{\mathrm{lin}}
=
\frac{\phi(\mathbf{q}_i)^\top\mS_i}
     {\phi(\mathbf{q}_i)^\top\mathbf{z}_i},
\quad
\mS_i
=
\mS_{i-1}
+
\phi(\mathbf{k}_i)\mathbf{v}_i^\top,
\quad
\mathbf{z}_i
=
\mathbf{z}_{i-1}
+
\phi(\mathbf{k}_i).
\label{eq:linear_attention_memory}
\end{equation}
}

Gated linear-attention models often use the identity feature map and omit the explicit normalizer, giving the fast-weight form $\mathbf{y}_i=\mathbf{q}_i^\top\mS_i$ with $\mS_i=\mS_{i-1}+\mathbf{k}_i\mathbf{v}_i^\top$. GDN starts from this recurrent memory and combines Mamba2-style decay with the DeltaNet residual write~\citep{yang2025gatedgdn}. In our key-by-value state orientation, one GDN direction is defined as follows.

A GDN direction projects each hidden token $x_i$ into query, key, value, decay, write-rate, and output-gate variables:
{\small
\begin{equation}
(\mathbf{q}_i,\mathbf{k}_i,\mathbf{v}_i,a_i,b_i,\mathbf{g}_i)=W_{\mathrm{gdn}}x_i .
    \label{eq:gdn_projection}
\end{equation}
}
After normalizing keys and scaling queries, the gated delta state is
{\small
\begin{equation}
\begin{aligned}
\bar{\mS}_i&=d_i\mS_{i-1},\quad
r_i=k_i^\top\bar{\mS}_i,\quad
w_i=\beta_i(v_i-r_i),\\
\mS_i&=\bar{\mS}_i+k_iw_i^\top,\quad
y_i=q_i^\top\mS_i .
\end{aligned}
\label{eq:gdn_recurrence}
\end{equation}
}
Here $d_i\in(0,1]$ decays old associations, $r_i$ reads the value already stored at the incoming key, and $\beta_i$ controls how much residual $\mathbf{v}_i-r_i$ is written. Thus GDN is not claimed to be exactly softmax attention: it is a recurrent replacement for the attention numerator, with gated memory clearing and targeted delta updates.

\paragraph{Prefix-Preserving Bidirectionality.}
A single GDN scan is causal in its scan order. A left-to-right scan
misses future target tokens, while reversing the whole sequence would
move prefix/control tokens to semantically wrong positions. Let $L_v$
denote the valid body length. DeltaFlow reverses only valid body tokens:
{\small
\begin{equation}
    \mathcal{R}_B(X;L_v)
    =
    (p_{1:P},x_{L_v:1},x_{L_v+1:L}),
    \qquad
    \mathcal{R}_B^2(X;L_v)=X .
    \label{eq:body_reverse_operator}
\end{equation}
}
Here $x_{L_v:1}=(x_{L_v},\ldots,x_1)$; prefix and padding positions are
not reversed. For conditional tasks, the source remains in the prefix
and only the target body is reversed.
This operator induces two prefix-preserving directional GDN views:
% {\small
% \begin{equation}
% \begin{aligned}
% \mathcal{G}^{\rightarrow}_{\ell}(X)
% &=\mathrm{GDN}_{\ell}(X),\\
% \mathcal{G}^{\leftarrow}_{\ell}(X)
% &=\mathcal{R}_B\!\bigl(
%   \mathrm{GDN}_{\ell}\!\bigl(\mathcal{R}_B(X;L_v)\bigr);L_v
% \bigr).
% \end{aligned}
% \label{eq:gdn_bidirectional_views}
% \end{equation}
% }
{\footnotesize
\begin{equation}
\mathcal{G}^{\rightarrow}_{\ell}(X)
=\mathrm{GDN}_{\ell}(X),
\
\mathcal{G}^{\leftarrow}_{\ell}(X)
=\mathcal{R}_B\!\left(
\mathrm{GDN}_{\ell}\!\left(\mathcal{R}_B(X;L_v)\right);L_v
\right).
\label{eq:gdn_bidirectional_views}
\end{equation}
}
The two directional scans share all parameters and differ only in body
order; their aligned outputs are averaged before a single shared output
gate and projection.

For a body position $i$, $\mathcal{G}^{\rightarrow}$ summarizes $(p_{1:P},x_{1:i})$, while $\mathcal{G}^{\leftarrow}$ summarizes $(p_{1:P},x_{L:i})$ after flipping the body output back to the original order. Together, the two views recover the left/right support of full attention while preserving the prefix order required by ELF.

\paragraph{DeltaFlow-A / Alternating-Scan.}
DeltaFlow-A applies the single-direction recurrence in Eqs.~\eqref{eq:scan_bigdn_schedule} once per GDN layer. Let $X^{(\ell)}=(p_1,\ldots,p_P,x_1^{(\ell)},\ldots,x_L^{(\ell)})$ be the input to GDN layer $\ell$. With an alternating direction schedule $\sigma_\ell\in\{\rightarrow,\leftarrow\}$ over GDN-layer ordinals, the bidirectional context is scheduled across layers:
{\small
\begin{equation}
Y_{\mathrm{scan}}^{(\ell)} =
\begin{cases}
\hblue{\mathcal{G}^{\rightarrow}_{\ell}(X^{(\ell)})},
& \sigma_\ell=\rightarrow,\\[2pt]
\horange{\mathcal{G}^{\leftarrow}_{\ell}(X^{(\ell)})},
& \sigma_\ell=\leftarrow .
\end{cases}
\label{eq:scan_bigdn_schedule}
\end{equation}
}
The block then applies the usual residual update $X^{(\ell+1)}=X^{(\ell)}+\mathrm{Proj}_{\ell}(Y_{\mathrm{scan}}^{(\ell)})$ plus the feed-forward sublayer. Hence each DeltaFlow-A layer has one recurrent scan and no in-layer bidirectional merge. Odd/even GDN layers exchange which body side is summarized; because every reverse-body layer applies $\mathcal{R}_B$ before returning to the residual stream, the next layer always receives tokens in the original prefix-plus-body order.

\paragraph{DeltaFlow-P / Parallel-Scan.}
DeltaFlow-P applies the same directional recurrence twice inside one GDN layer. Following the LION chunking template~\citep{afzal2026linear}, the two directional scans can be packed along the batch/chunk dimension, but the operator being merged is the GDN output rather than exact full linear attention. The forward branch computes the original-order view and the backward branch computes the reverse-body view:

{\small
\begin{equation}
\mathbf{Y}_{\ell}^{\mathrm{bi}}(X)
=
\frac{1}{2}
\left(
\hblue{\mathcal{G}^{\rightarrow}_{\ell}(X)}
+
\horange{\mathcal{G}^{\leftarrow}_{\ell}(X)}
\right).
\label{eq:gdn_merge}
\end{equation}
}

For target token $i$, the first term carries the prefix and left context $(p_{1:P},x_{1:i})$, while the second term carries the prefix and right context $(p_{1:P},x_{L:i})$. The factor $1/2$ keeps the merged response on the same scale as a single GDN output before the output gate and projection. Therefore DeltaFlow-A and DeltaFlow-P share the same prefix-preserving non-causal target, but place bidirectionality at different depths: across layers for DeltaFlow-A, and inside each GDN layer for DeltaFlow-P. Unless otherwise specified, we use DeltaFlow to refer to the full proposed method instantiated with Parallel-Scan, noise-adaptive memory control, and TSC.

\subsection{Noise-Adaptive Memory Control}

A fixed recurrent policy may be suboptimal across diffusion
times. For each executed scan direction $s$---$s=\sigma_\ell$
for DeltaFlow-A and $s\in\{\rightarrow,\leftarrow\}$ for
DeltaFlow-P---we add timestep-dependent shifts to the decay
and write-rate logits:
\begin{equation}
\begin{aligned}
\tilde a_{i,h}^{s}(t)
&=
a_{i,h}^{s}+u_h^a(e(t)),\\
\tilde b_{i,h}^{s}(t)
&=
b_{i,h}^{s}+u_h^b(e(t)),\\
d_{i,h}^{s}(t)
&=
\exp\!\left[
-\exp(A_h)
\operatorname{softplus}\!\left(
\tilde a_{i,h}^{s}(t)+c_h
\right)
\right],\\
\beta_{i,h}^{s}(t)
&=
\sigma\!\left(
\tilde b_{i,h}^{s}(t)
\right).
\end{aligned}
\label{eq:noise_adaptive_control}
\end{equation}
The projections $u_h^a$ and $u_h^b$ are shared across scan
directions and initialized to zero, recovering the
noise-independent update at initialization.

Using the original-order outputs defined above, let
$\bar y_{\ell,i}=Y_{\mathrm{scan},i}^{(\ell)}$ for DeltaFlow-A
and $\bar y_{\ell,i}=Y_{\ell,i}^{\mathrm{bi}}$ for DeltaFlow-P.
Both variants then apply one post-scan output gate and one
output projection:
\begin{equation}
o_{\ell,i}
=
W_o\!\left(
\operatorname{RMSNorm}(\bar y_{\ell,i})
\odot
\operatorname{SiLU}(g_{\ell,i})
\right).
\label{eq:output_gate}
\end{equation}

\subsection{Temporal State Consistency}

To stabilize training across diffusion times, we introduce
Temporal State Consistency (TSC), which aligns hidden states
at two paired points on the same shared-noise trajectory.
TSC is attached after complete backbone block
$b_{\mathrm{TSC}}=6$---the fifth GDN layer in
$[\mathrm{GDN},\mathrm{GDN},\mathrm{GDN},\mathrm{Attn}]
\times 3$---following its mixer residual and feed-forward
sublayer.

For clean latent $z^\star$ and shared noise $\epsilon$, we
construct
\begin{equation}
\begin{aligned}
z_t
&=
t z^\star+(1-t)\epsilon,\\
t^+
&=
\tfrac{1+t}{2},
\qquad
z_{t^+}
=
t^+z^\star+(1-t^+)\epsilon.
\end{aligned}
\label{eq:TSC_pair}
\end{equation}
Both passes use the same network and non-temporal
conditions; only the latent and timestep change. The cleaner
branch is treated as a stop-gradient target.

Let $h_i(z,t)$ denote the state of token $i$ at the TSC
attachment point. We normalize it as
\begin{equation}
\begin{aligned}
r_i(z,t)
&=
\operatorname{RMSNorm}\!\left(
h_i(z,t);10^{-6}
\right),\\
\hat r_i(z,t)
&=
\frac{r_i(z,t)}
{\max\!\left\{
\|r_i(z,t)\|_2,10^{-12}
\right\}}.
\end{aligned}
\label{eq:TSC_state}
\end{equation}
The auxiliary loss over valid body positions is
\begin{equation}
\mathcal{L}_{\mathrm{TSC}}
=
\frac{1}{|\Omega|}
\sum_{i\in\Omega}
\left\|
\hat r_i(z_t,t)
-
\operatorname{sg}\!\left[
\hat r_i(z_{t^+},t^+)
\right]
\right\|_2^2.
\label{eq:TSC_loss}
\end{equation}

TSC is applied only to the denoising branch:
\begin{equation}
\begin{aligned}
\mathcal{L}(m)
&=
\mathcal{L}_{\mathrm{base}}(m)
+
\mathbf{1}[m=\mathrm{denoise}]\,
\omega(p)\mathcal{L}_{\mathrm{TSC}},\\
\omega(p)
&=
0.025
\left[
1+
\cos\!\left(
\pi\,
\operatorname{clip}\!\left(
\frac{p-0.5}{0.4},0,1
\right)
\right)
\right].
\end{aligned}
\label{eq:TSC_objective}
\end{equation}
Here $\mathbf{1}[\cdot]$ is an indicator. The weight is
$0.05$ for $p\leq0.5$, decays to zero over
$0.5<p<0.9$, and remains zero thereafter.

\subsection{Hybrid Stack and Chunk Kernels}

DeltaFlow uses a 12-layer hybrid stack with the repeated ordering $[\mathrm{GDN},\mathrm{GDN},\mathrm{GDN},\mathrm{Attn}]\times3$, comprising nine bidirectional GDN blocks and three full non-causal attention blocks. The attention blocks provide exact bidirectional correction, while the GDN blocks supply efficient recurrent global memory. We implement the bidirectional GDN path with a chunk kernel that batches forward and backward body scans along the batch dimension, keeps prefix states shared, and reverses only body outputs after the scan.

\section{Experiments}
\label{sec:experiments}

\paragraph{Dataset and Evaluation.}
For unconditional generation, we train on the pre-tokenized OpenWebText (OWT) T5 split~\citep{gokaslan2019openwebtext}, containing approximately 9B tokens per epoch, with packed sequences of length $L=1024$. We generate 1,000 samples and report generative perplexity (Gen. PPL) under GPT-2 Large~\citep{radford2019language} and average unigram entropy. For conditional generation, we follow ELF on WMT14 German-to-English translation~\citep{bojar-etal-2014-findings} ($L=128$; 144M target tokens) and XSum~\citep{narayan-etal-2018-dont} summarization ($L=1088$; 6M target tokens), evaluated using BLEU~\citep{papineni-etal-2002-bleu} and ROUGE-1/ROUGE-2/ROUGE-L~\citep{lin-2004-rouge}, respectively. Conditional sequences are not packed.

\paragraph{Model.}
All models use frozen contextual embeddings from the released T5-small encoder~\citep{2020t5} (35M parameters; embedding dimension 512) and a 128-dimensional ELF bottleneck. We use the ELF-B configuration with 12 layers, hidden size 768, 12 heads, and dropout 0. We compare four denoisers: ELF-full, the original ELF-B reference built entirely
  with full non-causal attention; LION-S, a bidirectional selective-decay
  linear-attention baseline using a 3D1A hybrid stack; and our two GDN variants,
  DeltaFlow-A (\emph{Alternating-Scan}) and DeltaFlow-P
  (\emph{Parallel-Scan}).

\paragraph{Training and Inference.}
  OWT models are trained at global batch size 512 and effective learning rate
  0.002 using Muon~\citep{jordan2024muon} for hidden matrices and
  AdamW~\citep{loshchilov2018decoupled} for the remaining parameters. The
  denoising and decoding branches are sampled with probabilities 0.8 and 0.2.
  ELF-full and LION-S are trained for five epochs (45B tokens), whereas
  DeltaFlow models use four epochs (36B tokens). Unconditional evaluation uses
  1,000 length-1024 samples with a 32-step stochastic differential equation
  sampler (SDE32) and self-conditioning classifier-free guidance (CFG) scale 3;
  conditional evaluation uses a 64-step ordinary differential equation sampler
  (ODE64) with external-condition CFG.

% Additional optimization, fine-tuning, and implementation details are provided in Appendix~\ref{app:training_eval_details}.

\section{Results and Analysis}

\label{sec:results}

\begin{figure}[t]
\centering
\includegraphics[width=0.7\columnwidth]{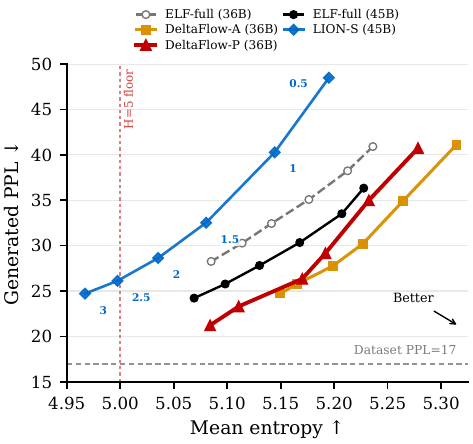}
% \captionof{figure}{CFG-control frontier over guidance scale for ELF-full, DeltaFlow-A and DeltaFlow-P under SDE 32-step evaluation. DeltaFlow-P shifts the PPL--entropy tradeoff toward lower PPL without leaving the ELF entropy band.}
\caption{
\textbf{SDE32 CFG frontiers on OpenWebText.}
Each curve shows how generated perplexity (PPL) and output diversity 
    (entropy) change as the guidance scale increases. 
}
%     OpenWebText SDE32 CFG frontiers for ELF-full, DeltaFlow-A,
% and DeltaFlow-P using PPL and entropy.
\label{fig:cfg_control}
\end{figure}
We organize the evaluation around four questions: whether DeltaFlow
improves the quality--diversity trade-off (RQ1), which components
contribute to its performance (RQ2), whether the backbone transfers
to conditional generation (RQ3), and whether it improves long-sequence
denoising efficiency (RQ4).

% \begin{table*}[t]
% \centering
% \caption{OpenWebText SDE32/CFG-3 comparison. PPL and entropy are
% reported with 95\% confidence intervals.}
% \label{tab:main_results_snapshot}
% \scriptsize
% \setlength{\tabcolsep}{3pt}
% \begin{tabular}{@{}lcccccc@{}}
% \toprule
% Method
% & Train data
% & PPL $\downarrow$
% & Entropy $\uparrow$
% & Dist-1/2/3 $\uparrow$
% & Rep-3 $\downarrow$
% & T5 JSD $\downarrow$ \\
% \midrule

% ELF-full
% & 45B
% & 24.218 [23.676, 24.647]
% & 5.069 [5.055, 5.083]
% & .0142/.2311/\textbf{.5615}
% & .1575
% & .0742 \\

% LION-S
% & 45B
% & 24.715 [24.136, 25.306]
% & 4.967 [4.945, 4.989]
% & .0131/.2279/.5611
% & .2004
% & .0731 \\

% \qcelllight{DeltaFlow-A}
% & \qcelllight{36B}
% & \qcelllight{24.724 [24.205, 25.275]}
% & \qcelllight{\textbf{5.149} [5.138, 5.161]}
% & \qcelllight{\textbf{.0146}/\textbf{.2314}/.5580}
% & \qcelllight{\textbf{.1408}}
% & \qcelllight{.0569} \\

% \qcell{DeltaFlow-P}
% & \qcell{36B}
% & \qcell{\textbf{21.228} [20.737, 21.721]}
% & \qcell{5.084 [5.067, 5.099]}
% & \qcell{.0140/.2253/.5491}
% & \qcell{.1564}
% & \qcell{\textbf{.0557}} \\

% \bottomrule
% \end{tabular}
% \end{table*}

\begin{table}[htbp]
\centering
\caption{OpenWebText SDE32/CFG-3 comparison. PPL and entropy are
reported with 95\% bootstrap confidence intervals over generated
samples from a single trained checkpoint.}
\label{tab:main_results_snapshot}

\scriptsize
\setlength{\tabcolsep}{1.2pt}

\begin{tabular}{@{}lcccccc@{}}
\toprule
Method
& \shortstack{Train\\Data}
& PPL $\downarrow$
& Entropy $\uparrow$
& \shortstack{Dist-1/2/3\\$\uparrow$}
& \shortstack{Rep-3\\$\downarrow$}
& \shortstack{T5 JSD\\$\downarrow$} \\
\midrule

ELF-full
& 45B
& \shortstack{24.218\\\mbox{[23.676,24.647]}}
& \shortstack{5.069\\\mbox{[5.055,5.083]}}
& \shortstack{.0142/.2311/\\\textbf{.5615}}
& .1575
& .0742 \\

LION-S
& 45B
& \shortstack{24.715\\\mbox{[24.136,25.306]}}
& \shortstack{4.967\\\mbox{[4.945,4.989]}}
& \shortstack{.0131/.2279/\\.5611}
& .2004
& .0731 \\

\qcelllight{DeltaFlow-A}
& \qcelllight{36B}
& \qcelllight{\shortstack{24.724\\\mbox{[24.205,25.275]}}}
& \qcelllight{\shortstack{\textbf{5.149}\\\mbox{[5.138,5.161]}}}
& \qcelllight{\shortstack{\textbf{.0146}/\textbf{.2314}/\\.5580}}
& \qcelllight{\textbf{.1408}}
& \qcelllight{.0569} \\

\qcell{DeltaFlow-P}
& \qcell{36B}
& \qcell{\shortstack{\textbf{21.228}\\\mbox{[20.737,21.721]}}}
& \qcell{\shortstack{5.084\\\mbox{[5.067,5.099]}}}
& \qcell{\shortstack{.0140/.2253/\\.5491}}
& \qcell{.1564}
& \qcell{\textbf{.0557}} \\

\bottomrule
\end{tabular}
\end{table}

% \begin{table*}[t]
% \centering
% \caption{WMT14 De-En and XSum transfer results for baselines,
% ELF-full, and DeltaFlow-P.}
% \label{tab:downstream_transfer_template}
% \footnotesize
% \begin{tabular}{@{}llcccc@{}}
% \toprule
% Method & Model size & De-En BLEU $\uparrow$ & XSum ROUGE-1 $\uparrow$ & XSum ROUGE-2 $\uparrow$ & XSum ROUGE-L $\uparrow$ \\
% \midrule
% AR & 99M & 25.2 & 30.5 $\pm$ 0.13 & 10.2 $\pm$ 0.11 & 24.4 $\pm$ 0.12 \\
% MDLM & 99M & 18.4 & 33.4 $\pm$ 0.11 & 11.6 $\pm$ 0.10 & 25.8 $\pm$ 0.10 \\
% Duo & 170M (+35M) & 21.3 & 31.4 $\pm$ 0.12 & 10.1 $\pm$ 0.10 & 25.0 $\pm$ 0.12 \\
% E2D2 & 99M & 24.8 & 28.4 $\pm$ 0.11 & 8.3 $\pm$ 0.09 & 22.0 $\pm$ 0.10 \\
% SeqDiffuSeq & -- & 21.3 & 19.3 & 1.7 & 14.1 \\
% CDCD & -- & 24.9 & -- & -- & -- \\
% \midrule
% ELF-full & 105M (+35M) & 26.94
% & 36.32 $\pm$ 0.43
% & 12.35 $\pm$ 0.36
% & 27.87 $\pm$ 0.41 \\

% % LION-S & 105M (+35M) & \emph{4.0} & \emph{12.0} & \emph{1.0} & \emph{9.3} \\

% % DeltaFlow-A & 105M (+35M) & \emph{18.8} & \emph{32.4} & \emph{9.5} & \emph{24.8} \\

% \qcell{DeltaFlow-P}
% & \qcell{110M (+35M)}
% & \qcell{24.16}
% & \qcell{35.51 $\pm$ 0.43}
% & \qcell{12.22 $\pm$ 0.37}
% & \qcell{27.79 $\pm$ 0.42} \\
% \bottomrule
% \end{tabular}
% \end{table*}

\paragraph{Main OpenWebText Results and CFG Frontier.}
Table~\ref{tab:main_results_snapshot} reports the main
SDE32/CFG-3 results, while Figure~\ref{fig:cfg_control}
compares the quality--diversity frontiers across guidance
scales. We additionally include an ELF-full checkpoint
trained on 36B tokens, matching the nominal training-token
exposure of DeltaFlow-A and DeltaFlow-P, while retaining
ELF-full (45B) as the fully trained reference. Following ELF,
we treat entropy below $5.0$ as a potentially degenerate
regime.

At CFG-3, DeltaFlow-A improves over the token-matched
ELF-full (36B) checkpoint and remains close to ELF-full
(45B) in PPL while producing higher entropy. DeltaFlow-P
achieves the strongest entropy-valid result, with a PPL of
$21.228$ and an entropy of $5.084$. Compared with the
fully trained ELF-full (45B), it reduces PPL from $24.218$
to $21.228$ while maintaining comparable entropy
($5.069$ versus $5.084$), corresponding to a $12.3\%$
reduction despite using 9B fewer training tokens. It also
outperforms the matched-exposure ELF-full (36B), indicating
that the improvement is not explained by additional
training-token exposure.

Across guidance scales, increasing CFG generally lowers
PPL but also reduces entropy. Within the entropy-valid
region, DeltaFlow-P consistently shifts the frontier toward
lower PPL relative to both ELF-full checkpoints, showing
that its advantage is not specific to CFG-3. In contrast,
DeltaFlow-A favors higher diversity with more modest PPL
gains. Overall, the matched-token comparison provides
stronger evidence that the improved trade-off arises from
the DeltaFlow backbone rather than unequal token exposure.
% \paragraph{Main OpenWebText Results.}
% Table~\ref{tab:main_results_snapshot} compares all models under the shared
% SDE32/CFG-3 protocol. Following ELF, we treat entropy below $5.0$ as a
% potentially degenerate regime. DeltaFlow-A remains close to ELF-full in
% PPL while producing higher entropy. DeltaFlow-P achieves the strongest
% valid result, reducing PPL from $24.218$ to $21.228$ while maintaining
% comparable entropy ($5.084$ versus $5.069$), corresponding to a $12.3\%$
% PPL reduction.

% \paragraph{CFG Frontier.}
% Figure~\ref{fig:cfg_control} shows that as the guidance scale
% increases, models generally achieve lower perplexity but also become more
% likely to enter low-entropy regions. DeltaFlow-A maintains relatively high
% entropy across the evaluated guidance scales, although its PPL improvement
% is limited. DeltaFlow-P shifts the valid
% PPL--entropy frontier toward lower perplexity across guidance scales,
% indicating that its improvement is not limited to the selected CFG-3
% operating point.

% \paragraph{Qualitative Generation Behavior.}
% Figure~\ref{fig:qualitative_examples} presents representative unconditional
% generations produced by DeltaFlow-P. The samples maintain basic syntactic
% coherence and demonstrate the ability to form locally consistent themes.

\label{sec:rq2}

 \begin{figure}[t]
      \centering
      \includegraphics[
          width=0.68\columnwidth
      ]
      {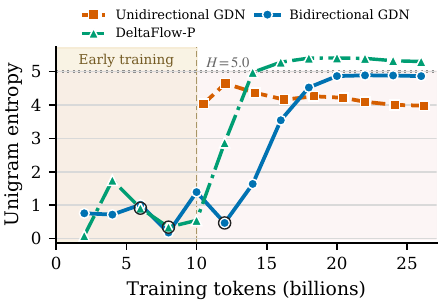}
      \caption{
          \textbf{Early-training entropy trajectories.}
      The bidirectional GDN approaches $H=5.0$, while the unidirectional GDN
      remains lower.
      }
      \label{fig:unidirectional-vs-bidirectional}
  \end{figure}

  \subsection{RQ2: Contribution of Individual Components}
\begin{figure}[t]
\centering
\includegraphics[width=\columnwidth]{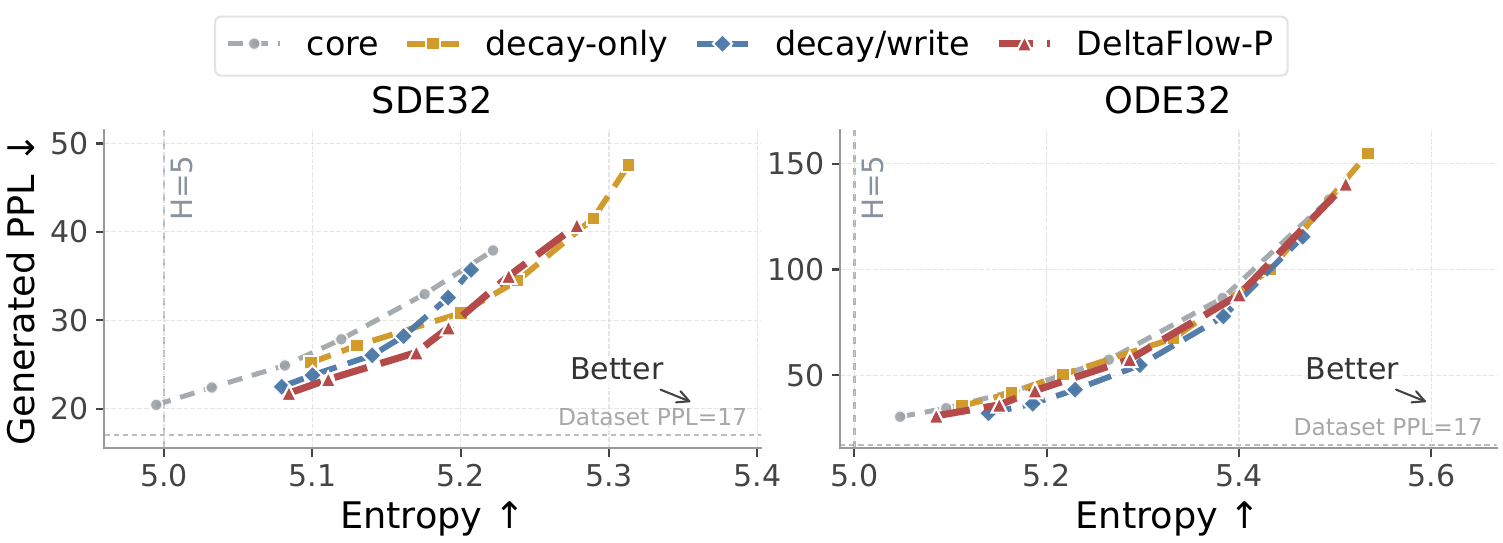}
% \caption{Component-ablation CFG frontiers under SDE32 and ODE32. Each line is a PPL--entropy sweep over CFG scale for one DeltaFlow component row. Dashed guides mark the $H=5$ entropy floor and the dataset PPL reference.}
\caption{\textbf{OpenWebText SDE32/ODE32 CFG frontiers for DeltaFlow-P ablations}, measured by generated perplexity (PPL) and average unigram entropy. Dashed lines indicate the entropy threshold ($H=5$) and dataset PPL. Bidirectional GDN lowers PPL but reduces diversity; noise-adaptive memory control restores a better balance, while TSC further improves the quality--diversity trade-off.}
    
\label{fig:component_ablation_frontier}
\end{figure}

\paragraph{Sampler Sensitivity and Component Ablation.}
As shown in
Figure~\ref{fig:unidirectional-vs-bidirectional}, the unidirectional GDN
suffers from entropy collapse and remains below the $H=5.0$ diversity
threshold, whereas bidirectional context aggregation progressively
recovers entropy. 
Figure~\ref{fig:component_ablation_frontier} shows that SDE32 generally
achieves lower PPL than ODE32 at comparable entropy, suggesting that the
gap primarily arises from sampling quality rather than output diversity.
We therefore use SDE32 for the main OpenWebText evaluation and retain
ODE32 as a complementary sampler stress test. The ablation frontiers in
Figure~\ref{fig:component_ablation_frontier} further show that the
bidirectional core provides the main PPL improvement, although it alone
still produces overly concentrated generations. Noise-adaptive decay
restores diversity, while joint decay/write control improves the
quality--diversity balance by suppressing unreliable updates at high
noise levels and enabling selective corrections at low noise levels.
Finally, TSC aligns representations across neighboring noise levels
during intermediate training. Together, these components enable the full
DeltaFlow model to achieve $21.228$ PPL with an entropy of $5.084$,
placing it above the $H=5.0$ diversity threshold.

% \subsection{RQ2: Contribution of Individual Components}
% \paragraph{Sampler Sensitivity.}
% Figure~\ref{fig:component_ablation_frontier} shows that SDE32 achieves
% lower PPL than ODE32 at comparable entropy, suggesting that the gap mainly
% arises from sampling quality rather than output diversity. We therefore use
% SDE32 for the main OpenWebText evaluation and ODE32 as a complementary
% sampler stress test.

% \paragraph{Component Ablation.}
% As shown in Figure~\ref{fig:unidirectional-vs-bidirectional}, the
% unidirectional GDN suffers from entropy collapse and remains below the
% $H=5.0$ diversity threshold, whereas bidirectional context aggregation
% progressively recovers entropy. The component-ablation frontiers in
% Figure~\ref{fig:component_ablation_frontier} further show that the
% bidirectional core provides the main PPL improvement, although it alone
% still produces overly concentrated generations. Noise-adaptive decay
% restores diversity, while joint decay/write control improves the
% quality--diversity balance by suppressing unreliable updates at high
% noise levels and enabling selective corrections at low noise levels.
% Finally, TSC aligns representations across neighboring noise levels
% during intermediate training. The resulting DeltaFlow model achieves
% $21.228$ PPL with an entropy of $5.084$, placing it above the $H=5.0$
% diversity threshold.

% Dashed horizontal and vertical guides mark the $H = 5$ entropy floor 
%     and the dataset PPL reference, respectively.
%     Each curve sweeps CFG guidance scale for one ablation configuration.
%     The bidirectional GDN core alone achieves low PPL but falls below the 
%     entropy threshold, indicating that bidirectional recurrence improves 
%     generation likelihood but can produce overly concentrated outputs.
%     Adding noise-adaptive decay restores entropy, while joint decay and 
%     write-rate control provides a better PPL-diversity balance.
%     Incorporating TSC yields the full DeltaFlow result of PPL 21.228 and 
%     entropy 5.084, improving the quality-diversity trade-off across 
%     both SDE32 and ODE32 samplers.

\subsection{RQ3: Transfer to Conditional Generation}
\label{sec:rq3}

\begin{table}[htbp]
\centering
\caption{Conditional transfer on WMT14 De--En and XSum.}
\label{tab:downstream_transfer_template}
\scriptsize
\setlength{\tabcolsep}{2.5pt}
\renewcommand{\arraystretch}{0.95}

\resizebox{\linewidth}{!}{%
\begin{tabular}{@{}lcccc@{}}
\toprule
& WMT14 & \multicolumn{3}{c}{XSum ROUGE} \\
\cmidrule(lr){2-2}
\cmidrule(lr){3-5}
Method & BLEU $\uparrow$
& ROUGE-1 $\uparrow$ & ROUGE-2 $\uparrow$ & ROUGE-L $\uparrow$ \\
\midrule

AR~\citep{vaswani2017attention} & 25.2
& $30.5{\pm}0.13$ & $10.2{\pm}0.11$ & $24.4{\pm}0.12$ \\
MDLM~\citep{sahoo2024mdlm} & 18.4
& $33.4{\pm}0.11$ & $11.6{\pm}0.10$ & $25.8{\pm}0.10$ \\
Duo~\citep{sahoo2025duo} & 21.3
& $31.4{\pm}0.12$ & $10.1{\pm}0.10$ & $25.0{\pm}0.12$ \\
E2D2~\citep{arriola2025e2d2} & 24.8
& $28.4{\pm}0.11$ & $8.3{\pm}0.09$ & $22.0{\pm}0.10$ \\
SeqDiffuSeq~\citep{yuan2024seqdiffuseq} & 21.3
& 19.3 & 1.7 & 14.1 \\
CDCD~\citep{dieleman2022cdcd} & 24.9
& -- & -- & -- \\
\midrule

ELF-full & \textbf{26.94}
& $\mathbf{36.32{\pm}0.43}$
& $\mathbf{12.35{\pm}0.36}$
& $27.87{\pm}0.41$ \\

 LION-S  & 24.12 & $32.21{\pm}0.42$ & $11.19{\pm}0.35$ & $25.40{\pm}0.40$ \\

\qcell{DeltaFlow-A}
& \qcell{24.55}
& \qcell{$35.32{\pm}0.45$}
& \qcell{$12.20{\pm}0.38$}
& \qcell{$\mathbf{28.10{\pm}0.44}$} \\

\qcell{DeltaFlow-P}
& \qcell{24.16}
& \qcell{$35.51{\pm}0.43$}
& \qcell{$12.22{\pm}0.37$}
& \qcell{$27.79{\pm}0.42$} \\

\bottomrule
\end{tabular}%
}
\end{table}

\paragraph{Machine Translation.}
  We evaluate conditional transfer on WMT14~\citep{bojar-etal-2014-findings}
  De--En using 64 source and 64 target tokens while retaining the same
  prefix-based conditioning interface as ELF. As shown in
  Table~\ref{tab:downstream_transfer_template}, DeltaFlow-P achieves a
  validation BLEU score of 24.16, while DeltaFlow-A reaches 24.55. These
  results compare with 26.94 for the full-attention ELF-full reference and
  24.12 for LION-S, our bidirectional selective-decay linear-attention
  baseline. DeltaFlow-P therefore matches LION-S on translation
  ($24.16$ versus $24.12$ BLEU), while remaining 2.78 BLEU behind ELF-full.
  This result shows that the hybrid recurrent backbone preserves conditional
  translation capability relative to the efficient linear-attention baseline,
  although a gap to full attention remains.

  \paragraph{Summarization.}
  On XSum, DeltaFlow-P achieves ROUGE-1, ROUGE-2, and ROUGE-L scores of
  $35.51$, $12.22$, and $27.79$, respectively, while DeltaFlow-A obtains
  $35.32$, $12.20$, and $28.10$. The corresponding scores are $36.32$,
  $12.35$, and $27.87$ for ELF-full and $32.21$, $11.19$, and $25.40$
  for LION-S. DeltaFlow-P improves over LION-S by 3.30 ROUGE-1,
  1.03 ROUGE-2, and 2.39 ROUGE-L points, substantially narrowing the gap
  to ELF-full. In particular, its differences from ELF-full are only
  0.13 on ROUGE-2 and 0.08 on ROUGE-L, while DeltaFlow-A obtains the
  highest ROUGE-L score in the table.

  Overall, both DeltaFlow variants consistently improve upon the LION-S
  baseline across the reported conditional-generation metrics. DeltaFlow-P
  nearly matches ELF-full on XSum and matches LION-S on WMT14 De--En,
  although its translation performance remains below the full-attention
  reference. These results support the transferability of the hybrid
  recurrent backbone across translation and summarization while leaving
  room for further improvement in translation accuracy.

\paragraph{Qualitative Conditional Examples.}
Figure~\ref{fig:qualitative_examples} presents representative examples of
translation and summarization. In the translation example, DeltaFlow-P
correctly converts the German input into an English sentence concerning
Obama's re-election. In the summarization example, the model captures the
central event---the UK government's planned assessment of paramilitary
organizations in Northern Ireland---and produces a concise summary.

% \begin{figure*}[t]
% \centering
% \includegraphics[width=0.95\textwidth]{Figures/elf_fig7ab_fig5c_triptych.pdf}
% \caption{ablation.}
% \label{fig:ablation}
% \end{figure*}

% gf
% \begin{table}[t]
% \centering
% % \caption{OpenWebText SDE32/CFG ablation of DeltaFlow-P, isolating
% % bidirectional GDN, decay/write control, and TSC, with PPL and entropy.}
% \caption{OpenWebText SDE32/CFG DeltaFlow-P ablation.}
% \label{tab:component_ablation}
% \small
% \begin{tabular}{@{}llrr@{}}
% \toprule
% Component & Representative row & PPL $\downarrow$ & Ent. $\uparrow$ \\
% \midrule
% Bidirectional core
% & Base only
% & 20.701
% & 4.984 \\

% \qcell{Decay control}
% & \qcell{Decay only}
% & \qcell{24.594}
% & \qcell{5.097} \\

% Decay/write control
% & Decay + write
% & 23.051
% & 5.064 \\

% \qcell{Full recipe}
% & \qcell{DeltaFlow}
% & \qcell{21.228}
% & \qcell{5.084} \\
% \bottomrule
% \end{tabular}
% \end{table}

\subsection{RQ4: Long-Sequence Computational Efficiency}
\label{sec:rq4}

 Table~\ref{tab:training_cost} summarizes schedule-equivalent training
  costs on four H100 GPUs. Under the reported training schedules,
  DeltaFlow-A and DeltaFlow-P require 173.2 and 216.2 GPU-hours for four
  epochs ($36$B nominal tokens), respectively, whereas LION-S and
  ELF-full require 223.9 and 229.2 GPU-hours for five epochs ($45$B
  tokens). These totals therefore describe the reported training
  budgets rather than an equal-token comparison. After normalization by nominal token exposure, DeltaFlow-A has the lowest cost at 4.81 GPU-h/B, followed by LION-S at 4.97 GPU-h/B. Although DeltaFlow-P has lower
  step throughput than ELF-full (0.353 versus 0.416 steps/s), it reaches
  a better PPL--entropy operating point within its shorter four-epoch
  schedule.

  Table~\ref{tab:h100_training_efficiency} reports denoiser-only
  efficiency on a single H100 while fixing the workload to $16{,}384$
  tokens per step. At a sequence length of $1$k, DeltaFlow-A and
  DeltaFlow-P provide $1.56\times$ and $1.16\times$ speedups over
  ELF-full, respectively; at $16$k, these speedups increase to
  $3.04\times$ and $2.72\times$. DeltaFlow-A retains memory usage
  comparable to ELF-full, whereas DeltaFlow-P consumes approximately
  21~GiB because it performs two scan directions. DeltaFlow-A therefore
  offers the stronger efficiency profile, while DeltaFlow-P trades
  additional computation and memory for improved generation quality.
  The $16$k measurements evaluate denoiser scaling only and should not
  be interpreted as evidence of end-to-end long-context generation
  quality.

\begin{table}[t]
    \centering
    \caption{
      Training budget and throughput-normalized nominal-cost projection
      under a common four-H100 configuration.
    }
    \label{tab:training_cost}
    \small
    \setlength{\tabcolsep}{4pt}
    \resizebox{\columnwidth}{!}{%
      \begin{tabular}{lrrrrr}
        \toprule
        Method
        & Params (M)
        & Tokens (B)
        & Step/s $\uparrow$
        & GPU-h/B $\downarrow$
        & Proj. GPU-h $\downarrow$ \\
        \midrule
        ELF-full
        & 104.6
        & 45
        & 0.416
        & 5.09
        & 229.2 \\

        LION-S
        & 104.7
        & 45
        & 0.426
        & 4.97
        & 223.9 \\

        DeltaFlow-A
        & 110.3
        & 36
        & 0.441
        & \textbf{4.81}
        & \textbf{173.2} \\

        DeltaFlow-P
        & 110.3
        & 36
        & 0.353
        & 6.01
        & 216.2 \\
        \bottomrule
      \end{tabular}%
    }
  \end{table}

\begin{figure}[!t]
\centering

% =========================================================
% Unconditional Generation
% =========================================================
\begin{tikzpicture}
\node[
    draw={rgb,255:red,155;green,203;blue,125},
    fill={rgb,255:red,244;green,250;blue,240},
    rounded corners=5pt,
    line width=0.6pt,
    inner sep=4pt,
    outer sep=0pt
] {%
\begin{minipage}{0.92\columnwidth}
\footnotesize
\raggedright
\setlength{\parindent}{0pt}
\setlength{\parskip}{0pt}

\textbf{\color[RGB]{79,142,53}Unconditional Generation}
\hfill
{\scriptsize\textbf{\color[RGB]{79,142,53}%
Gen. PPL: 21.74 \quad H: 5.08}}

\par
\vskip 2pt
{\color[RGB]{155,203,125}\hrule height 0.4pt}
\vskip 2.5pt
\noindent

For a long time, the first thing I did was to create my own
blog and text, and turn it all into my own piece of text.
That was a lot of my work, but the real way for me to do it
was to make it all into its own.
\end{minipage}%
};
\end{tikzpicture}

\par\smallskip

% =========================================================
% Translation
% =========================================================
\begin{tikzpicture}
\node[
    draw={rgb,255:red,145;green,169;blue,236},
    fill={rgb,255:red,244;green,247;blue,255},
    rounded corners=5pt,
    line width=0.6pt,
    inner sep=4pt,
    outer sep=0pt
] {%
\begin{minipage}{0.92\columnwidth}
\footnotesize
\raggedright
\setlength{\parindent}{0pt}
\setlength{\parskip}{0pt}

\textbf{\color[RGB]{49,88,183}Translation}
\hfill
{\scriptsize\textbf{\color[RGB]{49,88,183}BLEU: 30.5}}

\par
\vskip 2pt
{\color[RGB]{145,169,236}\hrule height 0.4pt}
\vskip 2.5pt
\noindent

\textbf{\color[RGB]{49,88,183}Context:}
Eine republikanische Strategie, um der Wiederwahl von Obama
entgegenzutreten.
\par\vspace{1pt}

\textbf{\color[RGB]{49,88,183}Reference:}
A Republican strategy to counter the re-election of Obama.
\par\vspace{1pt}

\textbf{\color[RGB]{49,88,183}Generated:}
A Republican strategy to deal with Obama's reelection.
\end{minipage}%
};
\end{tikzpicture}

\par\smallskip

% =========================================================
% Summarization
% =========================================================
\begin{tikzpicture}
\node[
    draw={rgb,255:red,233;green,161;blue,103},
    fill={rgb,255:red,255;green,247;blue,239},
    rounded corners=5pt,
    line width=0.6pt,
    inner sep=4pt,
    outer sep=0pt
] {%
\begin{minipage}{0.92\columnwidth}
\footnotesize
\raggedright
\setlength{\parindent}{0pt}
\setlength{\parskip}{0pt}

\textbf{\color[RGB]{200,100,22}Summarization}
\hfill
{\scriptsize\textbf{\color[RGB]{200,100,22}%
R-1: 66.7 \quad R-2: 32.0 \quad R-L: 66.7}}

\par
\vskip 2pt
{\color[RGB]{233,161,103}\hrule height 0.4pt}
\vskip 2.5pt
\noindent

\textbf{\color[RGB]{200,100,22}Context:}
The assessment, to be published in mid-October, will be used
to inform parties at Northern Ireland's political talks.
Northern Ireland Secretary Theresa Villiers said she would
also establish funding to tackle organised crime associated
with paramilitary groups.
\par\vspace{1pt}

\textbf{\color[RGB]{200,100,22}Reference:}
The government has commissioned an independent assessment
of paramilitary organisations in Northern Ireland.
\par\vspace{1pt}

\textbf{\color[RGB]{200,100,22}Generated:}
The UK government plans to carry out an assessment of
paramilitaries in Northern Ireland.
\end{minipage}%
};
\end{tikzpicture}

% \caption{Qualitative examples for unconditional generation,
% translation, and summarization.}
\caption{DeltaFlow examples on OpenWebText generation, WMT14 De-En
translation, and XSum summarization.}
\label{fig:qualitative_examples}
\end{figure}

\begin{table}[t]
\centering
% \caption{Denoiser-only H100 training efficiency at a fixed 16k token
% slots per step. Each mixer reports seconds per step and tokens per
% second in the upper row, and peak memory and speedup over ELF-full in
% the lower row. The additional TSC target forward is excluded from timing.}
% \caption{Denoiser-only H100 efficiency of ELF-full, LION-S,
% DeltaFlow-A, and DeltaFlow-P at fixed 16k token slots per step,
% reporting time, throughput, memory, and speedup; the additional TSC
% target forward is excluded.}
\caption{Denoiser-only H100 efficiency.}
\label{tab:h100_training_efficiency}

\footnotesize
\setlength{\tabcolsep}{2.2pt}
\renewcommand{\arraystretch}{0.98}

\resizebox{\columnwidth}{!}{%
\begin{tabular}{@{}c*{4}{cc}@{}}
\toprule
\multirow{2}{*}{Setting}
& \multicolumn{2}{c}{ELF-full}
& \multicolumn{2}{c}{LION-S}
& \multicolumn{2}{c}{DeltaFlow-A}
& \multicolumn{2}{c}{DeltaFlow-P} \\
\cmidrule(lr){2-3}
\cmidrule(lr){4-5}
\cmidrule(lr){6-7}
\cmidrule(l){8-9}
& \shortstack{Sec/step\\GiB}
& \shortstack{Tok/s\\Speedup}
& \shortstack{Sec/step\\GiB}
& \shortstack{Tok/s\\Speedup}
& \shortstack{Sec/step\\GiB}
& \shortstack{Tok/s\\Speedup}
& \shortstack{Sec/step\\GiB}
& \shortstack{Tok/s\\Speedup} \\
\midrule

\multirow{2}{*}{1k$\times$16}
& 0.189 & 86,571
& 0.140 & 117,211
& \textbf{0.121} & \textbf{134,899}
& 0.163 & 100,676 \\
& 15.07 & 1.00x
& 16.85 & 1.35x
& 15.16 & \textbf{1.56x}
& 21.11 & 1.16x \\
\addlinespace[1pt]

\multirow{2}{*}{2k$\times$8}
& 0.250 & 65,477
& 0.157 & 104,368
& \textbf{0.137} & \textbf{119,600}
& 0.178 & 92,258 \\
& 14.97 & 1.00x
& 16.74 & 1.59x
& 15.07 & \textbf{1.83x}
& 20.93 & 1.41x \\
\addlinespace[1pt]

\multirow{2}{*}{4k$\times$4}
& 0.375 & 43,715
& 0.191 & 85,602
& \textbf{0.170} & \textbf{96,623}
& 0.210 & 77,922 \\
& 14.92 & 1.00x
& 16.69 & 1.96x
& 15.03 & \textbf{2.21x}
& 20.87 & 1.78x \\
\addlinespace[1pt]

\multirow{2}{*}{8k$\times$2}
& 0.626 & 26,171
& 0.262 & 62,428
& \textbf{0.237} & \textbf{69,099}
& 0.278 & 58,945 \\
& 14.90 & 1.00x
& 16.67 & 2.39x
& 15.02 & \textbf{2.64x}
& 20.86 & 2.25x \\
\addlinespace[1pt]

\multirow{2}{*}{16k$\times$1}
& 1.132 & 14,473
& 0.407 & 40,286
& \textbf{0.372} & \textbf{44,044}
& 0.416 & 39,409 \\
& 14.90 & 1.00x
& 16.66 & 2.78x
& 15.04 & \textbf{3.04x}
& 20.87 & 2.72x \\

\bottomrule
\end{tabular}%
}
\end{table}

\section{Conclusion}

We introduced DeltaFlow, a bidirectional GDN backbone for efficient non-causal denoising in Embedded Language Flows. By combining prefix-preserving bidirectional recurrence, noise-adaptive memory control, and scheduled Temporal State Consistency, DeltaFlow improves the perplexity--entropy trade-off over full-attention ELF while achieving higher throughput on long sequences. These results show that DeltaFlow is a promising hybrid alternative that improves long-sequence denoiser throughput while retaining periodic full-attention correction.

% \section{Limitation}
% DeltaFlow remains a hybrid architecture with periodic full-attention layers and therefore does not achieve strictly linear end-to-end complexity. Its largest measured speedups occur beyond the training sequence length, where generation quality has not yet been evaluated. The reported quality confidence intervals are obtained from one trained checkpoint and do not capture training-run variability. Finally, the current conditional results show a remaining translation-quality gap relative to ELF-full.

\bibliography{aaai2027}

\end{document}